%% file: GoTube.tex
\def\year{2022}\relax
\documentclass[letterpaper]{article} 
\usepackage{aaai22}  
\usepackage{times}  
\usepackage{helvet}  
\usepackage{courier}  
\usepackage[hyphens]{url}  
\usepackage{graphicx} 
\usepackage{natbib}  
\usepackage{caption} 
\DeclareCaptionStyle{ruled}{labelfont=normalfont,labelsep=colon,strut=off} 
\usepackage[colorlinks=true, citecolor=black, linkcolor=black]{hyperref}
\usepackage{algorithm}
\usepackage{algorithmic}

\usepackage{newfloat}
\usepackage{listings}
\floatstyle{ruled}
\newfloat{listing}{tb}{lst}{}
\floatname{listing}{Listing}
\usepackage{array}
\newcommand{\PreserveBackslash}[1]{\let\temp=\\#1\let\\=\temp}
\newcolumntype{C}[1]{>{\PreserveBackslash\centering}p{#1}}
\usepackage{amsmath}
\usepackage{amsfonts}       
\usepackage{subcaption}
\usepackage[export]{adjustbox}
\usepackage{booktabs}       
\usepackage{url}            
\usepackage{multirow}
\usepackage{bbm}

\usepackage{color}

\newcommand{\R}{\mathbb{R}}

\newcommand{\rd}{\delta}
\newtheorem{definition}{Definition}
\newtheorem{lemma}{Lemma}

\newtheorem{theorem}{Theorem}

\newcommand{\calB}{\mathcal{B}}

\newcommand{\calP}{\bar{p}}
\newcommand{\calV}{\mathcal{V}}

\newcommand{\calS}{\mathcal{S}}

\DeclareMathOperator{\area}{Area}

\title{GoTube: Scalable Stochastic Verification of Continuous-Depth Models}
\author{
    Sophie Gruenbacher\,$^{1}$\thanks{$^{1}$TU Wien, $^{2}$IST Austria, $^{3}$CSAIL MIT, $^{4}$Stony Brook University. \newline Correspondence to: \texttt{sophie.gruenbacher@tuwien.ac.at}~~ \newline Code: \url{https://github.com/DatenVorsprung/GoTube}
    }~, Mathias Lechner\,$^{2}$, Ramin Hasani\,$^{3}$, \\
    \textbf{Daniela Rus\,$^{3}$, Thomas A. Henzinger\,$^{2}$, Scott A. Smolka\,$^{4}$, Radu Grosu\,$^{1}$}
}
\affiliations{

}

\usepackage{bibentry}

\begin{document}

\maketitle

\begin{abstract}
We introduce a new stochastic verification algorithm that formally quantifies the behavioral robustness of any time-continuous process formulated as a continuous-depth model. Our algorithm solves a set of global optimization (Go) problems over a given time horizon to construct a tight enclosure (Tube) of the set of all process executions starting from a ball of initial states. We call our algorithm GoTube. Through its construction, GoTube ensures that the bounding tube is conservative up to a desired probability and up to a desired tightness.
GoTube is implemented in JAX and optimized to scale to complex continuous-depth neural network models. Compared to advanced reachability analysis tools for time-continuous neural networks, GoTube does not accumulate overapproximation errors between time steps and avoids the infamous wrapping effect inherent in symbolic techniques. We show that GoTube substantially outperforms state-of-the-art verification tools in terms of the size of the initial ball, speed, time-horizon, task completion, and scalability on a large set of experiments.
GoTube is stable and sets the state-of-the-art in terms of its ability to scale to time horizons well beyond what has been previously possible.
\end{abstract}

\vspace*{-1ex}\section{Introduction}

\begin{figure}
\centering
\vspace{0mm}
\centering
\includegraphics[width=6.3cm]{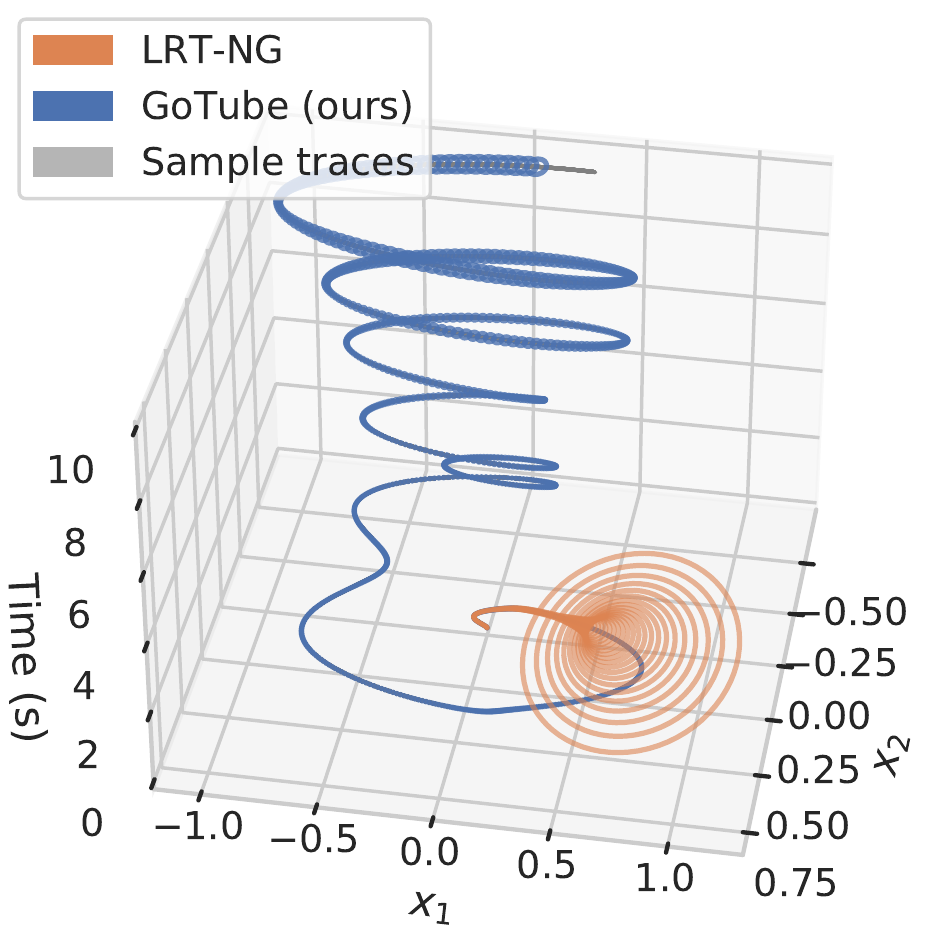}
\caption{Reachtubes of LRT-NG~\cite{gruenbacher2020lagrangian} and GoTube for a CT-RNN controlling CartPole-v1 environment. CAPD~\cite{capd} and Flow*~\cite{flowstar} failed.}
\label{fig:intro}
\end{figure}

The use of deep-learning systems powered by con\-tin\-u\-ous-depth models continues to grow, especially due to the revival of neural ordinary differential equations (Neural ODEs)~\cite{neuralODEs}. These models parametrize the derivative of the hidden states by a neural network. The resulting system of differential equations can perform strong function approximation and generative modeling. Ensuring their safety and robustness in any of these fronts is a major imperative, particularly in high-stakes decision-making applications such as medicine, automation, and finance.

A particularly appealing approach is to construct a tight overapproximation of the set of states reached over time according to the neural network's dynamics (a bounding tube) and provide deterministic 
or stochastic 
guarantees for the conservativeness of the tube's bounds.

Deterministic verification approaches ensure conservative bounds~\cite{flowstar,gowal2018effectiveness,mirman2018differentiable,bunel2020lagrangian,capd,gruenbacher2020lagrangian}, but often sacrifice speed and accuracy \cite{ehlers2017formal}, and thus scalability; see CAPD, Flow*, and LRT-NG in Fig.~\ref{fig:intro} and Fig.~\ref{fig:three graphs}. Stochastic methods, on the other hand, only ensure a weaker notion of conservativeness in the form of confidence intervals (stochastic bounds).  This, however, allows them to achieve much more accurate and faster verification algorithms that scale up to much larger dynamical system~\cite{probreach,gp,Gruenbacher2021verification}.

It was recently shown theoretically that stochastic verification approaches based on Lagrangian reachability (SLR) could provably guarantee confidence intervals for continuous-depth models~\cite{Gruenbacher2021verification}. The proposed theoretical framework suggests performing both stochastic global optimization and local differential optimization~\cite{stochGlobOptim, pontryagin2018mathematical}, and uses \emph{interval arithmetic} to symbolically bound the Lipschitz constant. Thus, it can construct a bounding ball of the reachable states at every time step, and over time, a tight bounding Tube. Although these theoretical results suggest an elegant way to avoid compounding errors, the SLR algorithm has not been implemented, so is this approach computationally tractable in practice?


We implemented the SLR algorithm as instructed in \cite{Gruenbacher2021verification}. We observed that even after resolving the first-occurring inefficient sampling and their vanishing gradient problems, the algorithm still blew up in time, even for low-dimensional benchmarks such as the Dubins Car. There are three fundamental algorithmic constraints of the symbolic techniques such as stochastic Lagrangian reachability that result in them being computationally intractable:
1)~the use of interval arithmetic for computing a conservative upper bound for the Lipschitz constant of the system fundamentally limits the scalability of reachability-based verification methods, 2)~the use of local gradient descent to search for local maxima in practice is more expensive than a simple local search, and 3)~the computational overhead due to the propagation of many initial states is high.



\begin{figure}[t]
\centering
\includegraphics[width=0.48\textwidth]{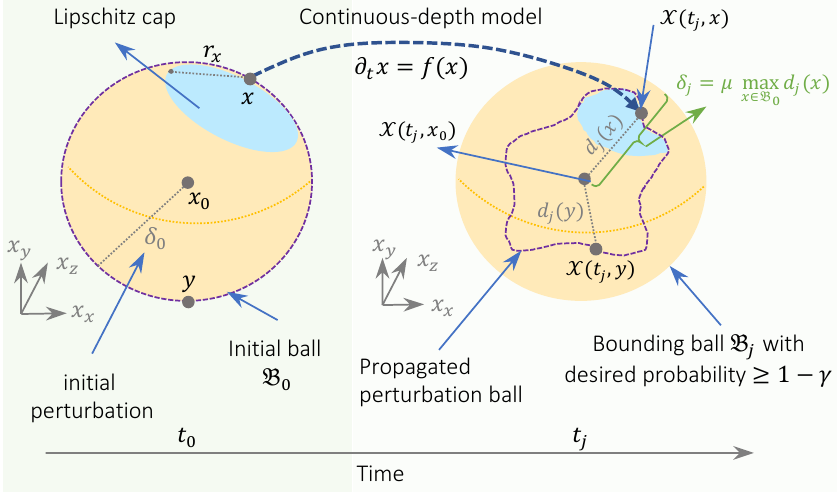}
\caption{GoTube in a nutshell. The center $x_0$ of ball $\calB_0\,{=}\,B(x_0, \rd_0)$, with $\rd_0$ the initial perturbation, and samples $x$ drawn uniformly from $\calB_0$'s surface, are numerically integrated in time to $\chi(t_j, x_0)$ and $\chi(t_j, x)$, respectively. The Lipschitz constant of $\chi(t_j, x)$ and their distance $d_j(x)$ to $\chi(t_j, x_0)$ are then used to compute Lipschitz caps around samples $x$, and the radius $\rd_j$ of bounding ball $\calB_j$ depending on the chosen tightness factor $\mu$. The ratio between the caps' surfaces and $\calB_0$'s surface are correlated to the desired confidence $1\,{-}\,\gamma$. }
\label{fig:gotube}
\vspace*{-4ex}
\end{figure}

In this work, we propose technical solutions for these fundamental issues and introduce a practical stochastic verification algorithm for continuous-time models. In particular, to tackle the first fundamental challenge introduced above, we develop a new theory that allows us to compute stochastic bounds for the Lipschitz constant in order to define a spherical cap around each sample, where the maximum perturbation is stochastically bounded (Lipschitz cap). As such, we are able to remove the conservative interval-based computation of the Lipschitz constant. Furthermore, we provide convergence guarantees for computing the upper bound of the confidence interval for the maximum perturbation at time $t_j$ with confidence level $1\,{-}\,\gamma$ and tube tightness $\mu$, using the estimation of the Lipschitz constant. This eliminates the dependence on the propagation horizon and considerably reduces computational complexity in the number of samples. We directly use this new Lipschitz constant computation framework instead of the costly interval arithmetic. 

We supply our global optimization scheme with a simple sampling process to propagate the initial states in parallel, according to the neural network's dynamics. This compensates for local differential optimization with additional samples. Our algorithm is called GoTube, as it solves a set of global optimization problems to construct a tight and computationally tractable enclosure (Tube) of all possible evolutions of the system for a given time horizon.

GoTube takes advantage of advanced automatic differential toolboxes such as JAX to perform highly parallel and tensorized operations to further enhance the runtime of the verification suite. On a large set of experiments with continuous-depth models, GoTube substantially outperforms state-of-the-art verification tools in terms of the size of the initial ball, speed, time-horizon, task completion, and scalability. We summarize the contributions of our paper as follows:

\begin{itemize}
    \item A novel and efficient theory for computing stochastic bounds for the Lipschitz constant of the system, which helps us achieve tight reachtubes for continuous-time dynamical systems.
    \item We prove convergence guarantees for the GoTube Algorithm, thus ensuring that the algorithm terminates in finite time even using stochastic Lipschitz caps around the samples instead of deterministic local balls.
    \item We perform a diverse set of experiments on continuous-time models with increasing complexity and demonstrate that GoTube considerably outperforms state-of-the-art verification tools.
\end{itemize}



\begin{table*}[t]
\small
\centering
\caption{Related work on the reachability analysis of continuous-time systems. Determ.= Deterministic. "No" indicates a stochastic method. Table content is partially reproduced from \cite{Gruenbacher2021verification}.}
\begin{adjustbox}{width=2\columnwidth}
\begin{tabular}{l|c|c|c|c}
\toprule
\textbf{Technique} & \textbf{Determ.} & \textbf{Parallel} & \textbf{wrapping} & \textbf{Arbitrary }\\
 & &  & \textbf{effect} & \textbf{Time-horizon}\\
\midrule
LRT \cite{Cyranka2017} with Infinitesimal strain theory & yes & no & yes & no \\
CAPD \cite{capd} implements Lohner algorithm & yes & no &  yes & no  \\
Flow-star \cite{flowstar} with Taylor models & yes & no  & yes & no  \\
$\delta$-reachability \cite{deltadecidable} with approximate satisfiability & yes & no & yes & no \\
C2E2 \cite{c2e2} with discrepancy functions & yes & no  & yes & no \\
LDFM \cite{fansimul} by simulation, matrix measures & yes & yes  & no & no \\
TIRA \cite{tira} with second-order sensitivity & yes & yes & no & no \\
Isabelle/HOL \cite{isabelle} with proof-assistant & yes & no  & yes & no \\
Breach \cite{breach,donze} by simulation & yes & yes  & no & no \\
PIRK \cite{pirk} with contraction bounds & yes & yes & no & no \\
HR \cite{hr} with hybridization & yes & no & yes & no \\
ProbReach \cite{probreach2} with $\delta$-reachability,  & no & no  & yes & no \\
VSPODE \cite{reliablecomput} using p-boxes & no &no  & yes & no \\
Gaussian process (GP) \cite{gp} & no & no & no & no \\
Stochastic Lagrangian reachability SLR \cite{Gruenbacher2021verification} & no & yes & no & no \\
\textbf{GoTube (Ours)} & no & yes & no & \textbf{yes} \\
\bottomrule
\end{tabular}
\end{adjustbox}
 \label{tab:related_works}
\end{table*}

\section{Related Work}
\textbf{Global Optimization.} Efficient local optimization methods such as gradient descent cannot be used for global optimization since such problems are typically non-convex. Thus, many advanced verification algorithms tend to use global optimization schemes \cite{NEURIPS2018_be53d253,bunel2020lagrangian}. Depending on the properties of the objective function, e.g. smoothness, various types of global optimization techniques exist. For instance, interval-based branch-and-bound (BaB) algorithms \cite{neumaier_2004,hansen2003global} work well on differentiable objectives up to a certain scale, which has recently been improved \cite{de2021improved}. There are also Lipschitz-global optimization methods for satisfying Lipschitz conditions \cite{Lipschitz,kvasov2013Lipschitz}. For example, a method for computing the Lipschitz constant of deep neural networks to assist with their robustness and verification analyses was recently proposed in~\cite{NEURIPS2019_95e1533e} and~\cite{bhowmick2021lipbab}. Additionally, there are evolutionary strategies for global optimization using the covariance matrix computation~\cite{cma,cmaes}. In our approach, for global optimization, we use random sampling and compute neighborhoods (Lipschitz caps) of the samples, where we have probabilistic knowledge about the values, such that we are able to correspondingly estimate the stochastic global optimum with high confidence.~\cite{stochGlobOptim}.

\noindent\textbf{Verification of Neural Networks.}
A large body of work tried to enhance the robustness of neural networks against adversarial examples \cite{goodfellow2014explaining}. There are efforts that show how to break the many defense mechanisms proposed \cite{athalye2018obfuscated,lechner2021adversarial}, until the arrival of methods for formally verifying robustness to adversarial attacks around neighborhoods of data \cite{henzinger2021scalable}. 
The majority of these complete verification algorithms for neural networks work on piece-wise linear structures of small-to-medium-size feedforward networks \cite{NEURIPS2019_246a3c55}. For instance, \cite{bunel2020branch} has recently introduced a BaB method that outperforms state-of-the-art verification methods \cite{katz2017reluplex,NEURIPS2020_f6c2a0c4}. A more scalable approach for rectified linear unit (ReLU) networks \cite{nair2010rectified} was recently proposed based on Lagrangian decomposition; this approach significantly improves the speed and tightness of the bounds \cite{de2021improved}. The proposed approach not only improves the tightness of the bounds but also performs a novel branching that matches the performance of the learning-based methods \cite{lu2020nueral} and outperforms state-of-the-art methods \cite{NEURIPS2018_d04863f1,singheth,bak2020improved,henriksen2020efficient}. While these verification approaches work well for feedforward networks with growing complexity, they are not suitable for recurrent and continuous neural network instances, which we address in this work. 

\noindent\textbf{Verification of Continuous-time Systems.} Reachability analysis is a verification approach that provides safety guarantees for a given continuous dynamical system \cite{gurung2019parallel,vinod2021stochastic}. Most dynamical systems in safety-critical applications are highly nonlinear and uncertain in nature \cite{lechner2020neural}. The uncertainty can be in the system's parameters \cite{sreach,probreach,reliablecomput}, or their initial state~\cite{reliablecomput,10.1145/3126508}. This is often handled by considering balls of a certain radius around them. Nonlinearity might be inherent in the system dynamics or due to 
discrete mode-jumps \cite{e71eb65e23844408b72fe95a84f88cb6}.
We provide a summary of methods developed for the reachability analysis of continuous-time ODEs in Table~\ref{tab:related_works}. 

A fundamental shortcoming of the majority of the methods described in Table~\ref{tab:related_works} is their lack of scalability while providing conservative bounds.
In this paper, we show that GoTube establishes the state-of-the-art for the verification of ODE-based systems in terms of speed, time-horizon, task completion, and scalability on a large set of experiments.



\section{Setup}
In this section, we introduce our notation, preliminary concepts, and definitions required to state and prove the stochastic bounds that GoTube guarantees for time-continuous process models.

\emph{Continuous-depth models.} These are a special case of non\-li\-near ordinary differential equations (ODEs), where the model is defined by the derivative of the unknown states $x$ computed by a vector-valued function $f\,{:}\,\R^n\,{\rightarrow}\,\R^n$, which is assumed to be Lipschitz-continuous and forward-complete:
\begin{align}\label{eq:ODE}
    \partial_t x = f(x),\quad x(t_0) \in \calB_0\,{=}\,B(x_0, \rd_0),
\end{align}
$\calB_0$ defines the initial ball (a region of initial states, whose radius quantifies the magnitude $\rd_0$ of a perturbation of its center $x_0$).
Time dependence can be incorporated by an additional variable $x$ with $\delta_t x = 1$. Thus this definition naturally extends to time-varying ODEs. Nonlinear ODEs do not have in general closed-form solutions, and therefore one can not compute symbolically the solution $\chi(t_j, x)$ for all $x\,{\in}\,\calB_0$. For a sequence of $k$ timesteps from time $t_0$ until time horizon $T$: $t_0\,{<}\dots{<}\,t_k = T$, we use numerical ODE solvers to compute $\chi(t_j, x)$ of the initial value problem (IVP) in Eq.~\eqref{eq:ODE} at time $t_j$ starting at different points $x(t_0)\,{=}\,x$. 

We extend this computation to the entire ball by numerically integrating the center $x_0$ and a set of points $x\,{\in}\,\calV$, uniformly sampled from the surface of the ball, and using this information to compute stochastic upper bounds for the possible evolutions of the system.
We define the bounding ball and bounding tube as follows:
\begin{definition}[Bounding Ball]
     Given an initial ball $\calB_0\,{=}\,B(x_0, \rd_0)$, we call $\calB_j = B(\chi(t_j, x_0), \rd_{j}(\calB_0))$ a {\em bounding ball} at time $t_j$, if it stochastically bounds the reachable states $x$ at time $t_j$ for all initial points around $x_0$ having the maximal initial perturbation $\rd_0$.
\end{definition}
%
As we do not only want to bound the perturbation at one specific time, but on a time series, we define:
\begin{definition}[Bounding Tube]
     Given an initial ball $\calB_0 = B(x_0, \rd_0)$ and bounding balls for $t_0\,{<}\dots{<}\,t_k\,{=}\,T$, we call the series of bounding balls $\calB_1, \calB_2, \dots, \calB_k$ a \emph{bounding tube}.
\end{definition}
\emph{Maximum perturbation at time $t_j$}. To compute a bounding tube, we have to compute at every timestep $t_j$ the maximum perturbation $\rd_j$, which is defined as a solution of the optimization problem:
\begin{align}\label{eq:optim}
    \rd_j \ge \max_{x\in\calB_0}\|\chi(t_j, x) - \chi(t_j, x_0)\| = \max_{x\in\calB_0} d(t_j, x),
\end{align}
where $d_j(x)=d(t_j,x)$ denotes the {\em distance} at time $t_j$, if the initial center $x_0$ is known from the context. As stated in~\cite{Gruenbacher2021verification}, the radius at time $t_j$ can be over-approximated by solving a global optimization problem on the surface of the initial ball $\calB_0$: as we require Lipschitz-continuity and forward-completeness of the ODE in Eq.~\eqref{eq:ODE}, the map $x \mapsto \chi(t_j,x)$ is a homeomorphism and commutes with closure and interior operators. In particular, the image of the boundary of the set $\calB_0$ is equal to the boundary of the image $\chi(t_j,\calB_0)$. Thus, Eq.~\eqref{eq:optim} has its optimum on the surface of the initial ball $\calB_0^S = \textrm{surface}(\calB_0)$, and we will only consider points on the surface.


\section{Main Results}
Our GoTube algorithm and its theory solve fundamental scalability problems of related works (see Table~\ref{tab:related_works}) by replacing interval arithmetic used to compute deterministic caps with stochastic Lipschitz caps. This enables us to verify continuous-depth models up to an arbitrary time-horizon, a capability beyond what was achievable before.

To be able to do that, we formulated Theorems on: 1)~How to choose the radius of a Lipschitz cap using stochastic bounds of local Lipschitz constants of the samples together with the expected difference quotients. 2)~Convergence guarantees using these new stochastic caps, as they are used by GoTube to compute the probability of $\rd_j$ being an upper bound of the biggest perturbation.
In addition, we implemented tensorization and substantially increased the number of random samples, thus being able to remove the dependence on the propagation-horizon of the gradient descent and increasing the computation speed to be able to deal with continuous-depth models.

\begin{algorithm}[t]
    \caption{GoTube}
    \label{algorithm:GoTube}
    \begin{algorithmic}[1]
    \REQUIRE initial ball $\calB_0 = B(x_0, \rd_0)$, time horizon T, sequence of timesteps $t_j$ ($t_0<\dots< t_k=T$), error tolerance $\mu\,{>}\,1$, confidence level $\gamma\,{\in}\, (0,1)$, batch size $b$, distance function $d$
    \vspace*{2mm}
    \STATE $\calV\leftarrow\{\}$ \quad(list of visited random points)
    \STATE \textbf{sample batch} $x^B \in \calB_0^S$
    \FOR{$(j=1; j\le k; j=j+1)$}\label{line:for loop Reachsets}
        \STATE $\calP\leftarrow 0$
        \WHILE{$\calP < 1 - \gamma$}
            \STATE $\calV\leftarrow \calV \cup \{x^B\}$
            \STATE $x_j \leftarrow \chi(t_j, x_0)$\quad (integrate initial center point)
            \STATE $\bar{m}_{j,\calV} \leftarrow \max_{x\in\calV} d(t_j, x)$
            \STATE \textbf{compute} local Lipschitz constants $\lambda_x$ for $x\in\calV$
            \STATE \textbf{compute} expected local difference quotient $\Delta\lambda_{x,\calV}$ for $x\in\calV$
            \STATE \textbf{compute} cap radii $r_x(\lambda_x, \Delta\lambda_{x,\calV})$ (Thm.~\ref{thm:lipschitz cap}) for $x\in\calV$\label{line:increase radii}
            \STATE $\calS\leftarrow \bigcup_{x\in\calV} B(x,r_x)^S$ \quad (total covered area)
            \STATE $\calP \leftarrow \Pr(\mu \cdot \bar{m}_{j,\calV} \ge m^\star)$
            \STATE \textbf{sample batch} $x^B \in \calB_0$
        \ENDWHILE
        \STATE $\rd_j\leftarrow \mu\cdot\bar{m}_{j,\calV}$
        \STATE $\calB_j\leftarrow B(x_j, \rd_j)$
    \ENDFOR
    \RETURN $(\calB_1,\dots,\calB_k)$
    \end{algorithmic}
\end{algorithm}

We start by describing the GoTube Algorithm. This facilitates the comprehension of the different computation and theory steps. Given a continuous-depth model as in Eq.~\eqref{eq:ODE}, an initial ball $\calB_0$ defined by a center point $x_0$ and the maximum initial perturbation $\rd_0$, a time horizon $T$ with a sequence of timesteps $t_j~(t_0\,{<}\dots{<}\,t_k=T)$, a confidence level $\gamma\,{\in}\,(0,1)$, a tightness factor $\mu\,{>}\,1$, a batch size $b$, and a distance function $d$. The output of the GoTube algorithm is a bounding tube that stochastically over-approximates at most by $\mu$ the propagated initial perturbation from the center $x_0$ with a probability higher than $1\,{-}\,\gamma$.

GoTube starts by sampling a batch (tensor) $x^B\in\calB_0^S$. It then iterates for the $k$ steps of the time horizon $T$ the following. After initializing the probability ensured to zero, and the visited states to the empty set, it loops until it reaches the desired confidence (probability) $1\,{-}\,\gamma$, by increasingly taking additional batches. In each iteration, it integrates the center and the already available samples from their previous time step and the possibly new batches from their initial state (for simplicity, the pseudocode does not make this distinction explicit). GoTube then computes the maximum distance from the integrated samples to the integrated center, their local Lipschitz constant
according to the variational equation of Eq.~\eqref{eq:ODE}. Based on this information GoTube then computes the mean Lipschitz statistics and the cap radii accordingly. The total surface of the caps is then employed to compute and update the achieved confidence (probability). Once the desired confidence is achieved, GoTube exits the inner loop and computes the bounding ball in terms of its center and radius, which is given by tightness factor $\mu$ times the maximum distance $\bar{m}_{j,\calV}$. After exiting the outer loop, GoTube returns the bounding tube.
\begin{definition}[Lipschitz Cap]
    Let $\calV$ be the set of all sampled points, $x\in\calV$ be a sample point on the surface of the initial ball, $\bar{m}_{j,\calV} = \max_{x\in\calV} d_j(x)$ be the sample maximum and $B(x,r_x)^S = B(x,r_x)\cap\calB_0^S$ be a spherical cap around that point. We call the cap $B(x,r_x)^S$ a $\gamma,t_j$-Lipschitz cap, if it holds that $\Pr\left(d_j(y) \le \mu\cdot \bar{m}_{j,\calV}\right)\ge 1-\gamma$ for all $y\in B(x,r_x)^S$.
\end{definition}
Lipschitz caps around the samples are a stochastic version of local balls around samples, commonly used to cover state space. Intuitively, the points within a cap do not have to be explored. The difference with Lipschitz caps is, that we stochastically bound the values inside that space and develop a theory to enable us to calculate a probability of having found an upper bound of the true maximum $m_j^\star = d_j(x_j^\star) = \max_{x\in\calB_0} d_j(x)$ of the optimization problem in Eq.~\eqref{eq:optim}.
Our objective is to avoid the usage of interval arithmetic for computing the Lipschitz constant, as it impedes scaling up to continuous depth models. Instead, we define stochastic bounds on the Lipschitz constant to set the radius $r_x$ of the Lipschitz caps, such that $\mu\cdot\bar{m}_{j,\calV}$ is a $\gamma$-stochastic upper bound for all distances $d_j(y)$ at time $t_j$ from values inside the ball $B(x, r_x)^S$.

\begin{theorem}[Radius of Stochastic Lipschitz Caps]\label{thm:lipschitz cap}
    Given a continuous-depth model $f$ from Eq.~\eqref{eq:ODE}, $\gamma\,{\in}\,(0,1)$, $\mu\,{>}\,1$, target time $t_j$,
    the set of all sampled points $\calV$, the number of sampled points $N\,{=}\,|\calV|$, the sample maximum $\bar{m}_{j,\calV}\,{=}\,\max_{x\in\calV} d_j(x)$, the IVP solutions $\chi(t_j,x)$, and the corresponding stretching factors $\lambda_x\,{=}\,\|\partial_x\chi(t_j,x)\|$ for all $x\,{\in}\,\calV$. Let us define $\hat{\gamma} = 1-\sqrt{1-\gamma}$.
    Let $\Delta\lambda_{\calV}$ be the $\sqrt{1-\gamma}$-quantile of a stochastic lower bound $F_{L,\hat{\gamma}}$ as defined by Lemma 1 in the Appendix:
    \begin{align}\label{eq:expected difference quotient}
        \Delta\lambda_{\calV}(\gamma) = F_{L,\hat{\gamma}}^{-1}(\sqrt{1-\gamma}),
    \end{align}
    Let $r_x$ be defined as:
    \begin{align}\label{eq:cap radius}
        r_{x} = \frac{\left(-\lambda_x + \sqrt{\lambda_x^2 + 4\cdot\Delta\lambda_{x,\calV}\cdot(\mu\cdot\bar{m}_{j,\calV}-d_j(x))}\right)}{2\cdot\Delta\lambda_{x,\calV}},
    \end{align}
    then it holds that:
    \begin{align}\label{eq:cap probability}
        \Pr\left(d_j(y) \le \mu\cdot \bar{m}_{j,\calV}\right)\ge 1-\gamma\quad \forall y\in B(x,r_x)^S,
    \end{align}
    and thus that $B(x, r_x)^S$ is a $\gamma, t_j$-Lipschitz cap.
\end{theorem}
The full proof is provided in the Appendix. \emph{Proof sketch:} As $\Delta\lambda_{x,\calV}$ is the $\sqrt{1-\gamma}$-quantile of $\max_{x,y}|\lambda_x-\lambda_y|/\|x-y\|$, it holds that $\Pr(\lambda_y \le \lambda_x + \Delta\lambda_{x,\calV} \cdot \|x-y\|)\ge 1-\gamma$. Therefore Eq.~\eqref{eq:cap radius} follows by solving the following equation: $(\mu\cdot\bar{m}_{j,\calV}-d_j(x))=\lambda_x r_x + \Delta\lambda_{x,\calV} r_x^2$.

Using conditional probability, we are able to state that the convergence guarantee holds for the GoTube Algorithm, thus ensuring that the Algorithm terminates in finite time even using stochastic Lipschitz caps around the samples instead of deterministic local balls.

\begin{theorem}[Convergence via Lipschitz Caps]\label{thm:stochastic guarantee} Given the tightness factor $\mu > 1$, the set of all sampled points $\calV$ and the sample maximum $\bar{m}_{j,\calV} = \max_{x\in\calV} d_j(x)$. Let the initial ball maximum be defined by $m^\star_j=\max_{x\in\calB_0} d_j(x)$. Then:
    \begin{align}\label{eq:stochastic guarantee}
        \hspace*{-2ex}\forall\gamma\in(0,1),\exists N\in\mathbb{N}\textrm{ s.t. }
        \Pr(\mu\cdot\bar{m}_{j,\calV}\ge m^\star_j) \ge 1-\gamma
    \end{align}
    where $N=|\calV|$ is the number of sampled points.
\end{theorem}
The full proof is provided in the Appendix. \emph{Proof sketch:} Let $x^\star_j$ be a point such that $d_j(x^\star_j) = m^\star_j$. Given $\gamma\in(0,1)$ and cap radii $r_x$, we expand the convergence guarantee from deterministic local balls to stochastic Lipschitz caps. For local balls it holds that $\exists N \in\mathbb{N}\colon\Pr(\exists x\in\calV\colon\,B(x,r_x)^S\owns x^\star_j)\ge\sqrt{1-\gamma}$. Using a set of sampled points $\calV$ with cardinality $N$ and using $1-\sqrt{1-\gamma}$ instead of $\gamma$
in Eq.~\eqref{eq:expected difference quotient} 
and Theorem~\ref{thm:lipschitz cap}, the resulting probability is larger than $\sqrt{1-\gamma}$. From the definition of a Lipschitz cap it follows that $\Pr( d_j(x^\star) \le \mu\cdot\bar{m}_{j,\calV} |\exists x\in\calV\colon B(x,r_x)^S\owns x^\star)\ge \sqrt{1-\gamma}$. For any sets $A, B$ it holds that $\Pr(A)\ge \Pr(A\cap B) = \Pr(A | B) \cdot \Pr(B)$, thus we multiply both probabilities and therefore Eq.~\eqref{eq:stochastic guarantee} holds.

\begin{figure*}[t]
     \centering
     \includegraphics[width=0.7\textwidth]{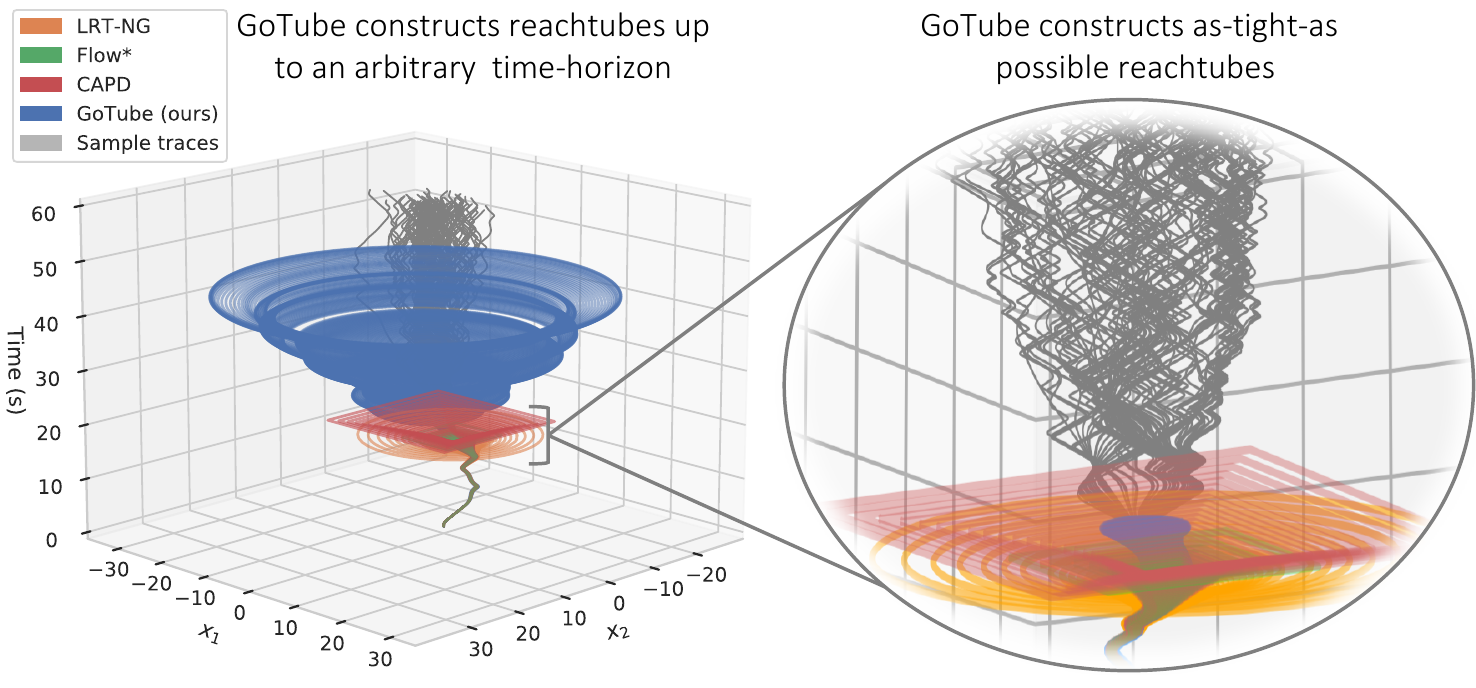}
        \caption{Visualization of the reachtubes constructed for the Dubin's car model with various reachability methods. While the tubes computed by existing methods (LRT-NG, Flow* and CAPD) explode at $t\,{\approx}\,20s$ (this moment is shown on the right side of the figure) due to the accumulation of over-approximation errors (the infamous wrapping effect), GoTube can keep tight bounds beyond $t\,{>}\,40s$ for a 99\% confidence level (using 20000 samples, $\mu=1.1$ and runtime of one hour). Note also the chaotic nature of 100 executions. }
        \label{fig:three graphs}
\end{figure*}

\section{Experimental Evaluation}
We perform a diverse set of experiments with GoTube to evaluate its performance and identify its characteristics and limits in verifying continuous-time systems with increasing complexity. We run our evaluations on a standard workstation machine setup (12 vCPUs, 64GB memory) equipped with a single GPU for a per-run timeout of 1 hour (except for runtimes reported in Figure \ref{fig:pareto}).

\subsection{On the volume of the bounding balls with GoTube}
Our first experimental evaluation concerns the overapproximation errors of the constructed bounding tubes.
An ideal reachability tool should be able to output an as tight as possible tube that encloses the system's executions. Consequently, as our comparison metric, we will report the average volume of the bounding balls, with less volume is better.
We use the benchmarks and settings of \cite{gruenbacher2020lagrangian} (same radii, time horizons, and models) as the basis of our evaluation. In particular, we compare GoTube to the deterministic, state-of-the-art reachability tools LRT-NG, Flow*, CAPD, and LRT. We measure the volume of GoTube's balls at the confidence levels of 90\% and 99\%, using $\mu=1.1$ as the tightness factor (in the third experiment we will talk in more detail about the trade-off between tightness and runtime).

The results are shown in Table \ref{tab:volume}. For the first five benchmarks, which are classical dynamical systems, we use the small time horizons $T$ and small initial radii $\rd_0$, which the other tools could handle. GoTube, with 99\% confidence, achieves a competitive performance to the other tools, coming out on top in 3 out of 5 benchmarks - using $\mu=1.1$ as the tightness bound. Intuitively this means, we are confident that the overapproximation includes all executions with a confidence level $1-\lambda$, but this overapproximation might not be as tight as desired. GoTube is able to achieve any desired tightness by reducing $\mu$ and increasing the runtime. The specific reachtubes and the chaotic nature of hundred executions of Dubin's car are shown in Figure \ref{fig:three graphs}. As one can see, the GoTube reachtube extends to a much longer time horizon, which we fixed at 40s. All other tools blew up before 20s. For the two problems involving neural networks, GoTube produces significantly tighter reachtubes.  

\begin{table}[t]
    \centering
    \caption{Comparison of GoTube (using tightness bound $\mu=1.1$) to existing reachability methods. The first five benchmarks concern classical dynamical systems, whereas the two bottom rows correspond to time-continuous RNN models (LTC= liquid time-constant networks) in a closed feedback loop with an RL environment \cite{Hasani2021liquid,vorbach2021causal}. The numbers show the volume of the constructed tube. Lower is better; best number in bold.    }
    \begin{adjustbox}{width=1\columnwidth}
    \begin{tabular}{l|llll|ll}
    \toprule \multirow{2}{*}{Benchmark} & \multirow{2}{*}{LRT-NG} &\multirow{2}{*}{Flow*}  &\multirow{2}{*}{CAPD}  & \multirow{2}{*}{LRT} & \multicolumn{2}{c}{GoTube}\\
         &    &   &  &  & (90\%) & (99\%)\\
       \midrule
       Brusselator  & 1.5e-4 & 9.8e-5  & 3.6e-4 & 6.1e-4 & 8.6e-5 & \textbf{8.6e-5} \\
       Van Der Pol &  4.2e-4 & \textbf{3.5e-4} & 1.5e-3  &  3.5e-4 & 3.5e-4 &  \textbf{3.5e-4} \\
       Robotarm & \textbf{7.9e-11} & 8.7e-10  & 1.1e-9 & Fail &  2.5e-10 & 2.5e-10 \\
       Dubins Car & 0.131  & 4.5e-2 & 0.1181 & 385 &  2.5e-2 & \textbf{2.6e-2} \\
       Cardiac Cell &  \textbf{3.7e-9} & 1.5e-8 & 4.4e-8  & 3.2e-8 &   4.2e-8 &  4.3e-8 \\
       \midrule
       CartPole-v1+LTC  & 4.49e-33  & Fail & Fail & Fail & 2.6e-37 &  \textbf{4.9e-37}  \\
       CartPole-v1+CTRNN  & 3.9e-27  & Fail & Fail & Fail &  9.9e-34 &  \textbf{1.2e-33}   \\
      \bottomrule
    \end{tabular}
    \end{adjustbox}
    \label{tab:volume}
\end{table}

\subsection{GoTube provides safety bounds up an arbitrary time horizon} 
In our second experiment, we evaluate for how long GoTube and existing methods can construct a reachtube before exploding due to overapproximation errors. To do so, we extend the benchmark setup by increasing the time horizon for which the tube should be constructed, use tightness bound $\mu=1.1$ and set a 95\% confidence level, that is, probability of being conservative. 
\begin{table}[ht]
\small
    \centering
    \caption{Results of the extended benchmark by longer time horizons. The numbers show the volume of the constructed tube, ``Blowup'' indicates that the method produced \texttt{Inf} or \texttt{NaN} values due to a blowup. Lower is better; the best method is shown in bold.}
    \begin{tabular}{l|cc|cc}\toprule
         Benchmark  & \multicolumn{2}{c|}{CartPole-v1+CTRNN} & \multicolumn{2}{c}{CartPole-v1+LTC} \\
         Time horizon  & 1s & 10s & 0.35s & 10s \\\midrule
         LRT & Blowup & Blowup & Blowup & Blowup \\
         CAPD & Blowup & Blowup & Blowup & Blowup \\
         Flow* & Blowup & Blowup & Blowup & Blowup\\
         LRT-NG & 3.9e-27 & Blowup & 4.5e-33 & Blowup\\
         GoTube (ours) & \textbf{8.8e-34} & \textbf{1.1e-19} & \textbf{4.9e-37} & \textbf{8.7e-21} \\\bottomrule
         \end{tabular}
    \label{tab:time}
\end{table}

The results in Table~\ref{tab:time} demonstrate that GoTube produces significantly longer reachtubes than all considered state-of-the-art approaches, without suffering from severe overapproximation errors. Particularly, Figure~\ref{fig:intro} visualizes the difference to the existing methods and overapproximation margins for two example dimensions of the CartPole-v1 environment and its CT-RNN controller.

\begin{figure*}[t]
    \centering
    \includegraphics[width=0.8\textwidth]{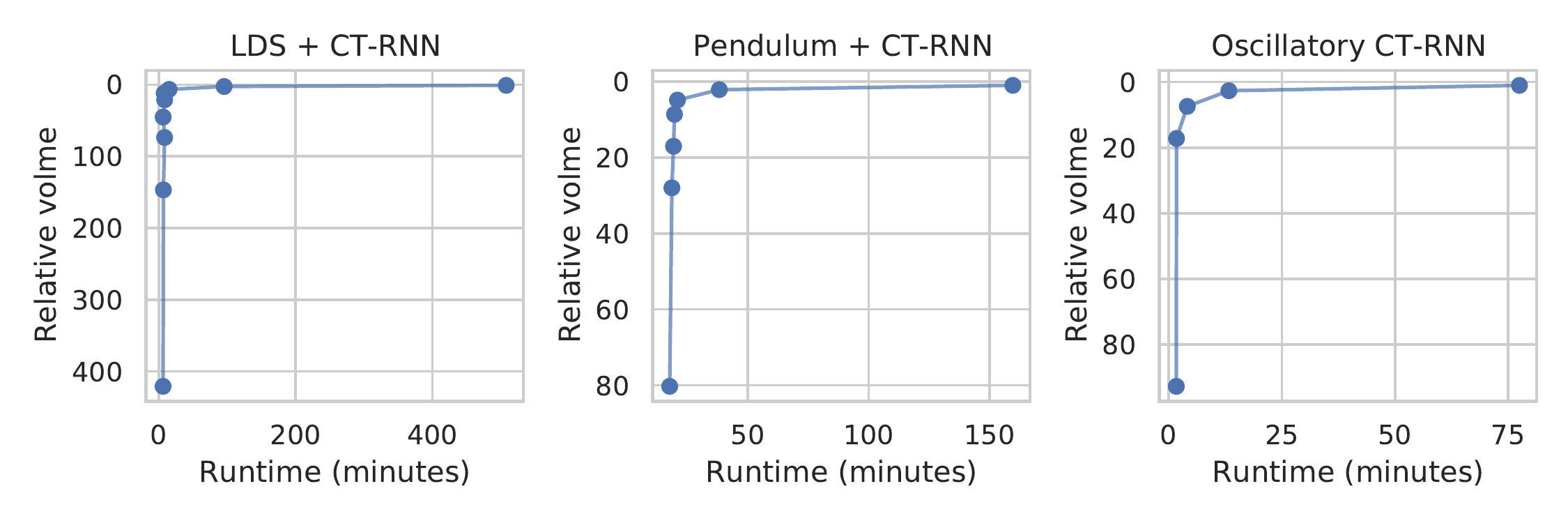}
    \caption{GoTube's runtime (x-axis) and volume size (y-axis) as a function of the tightness factor $\mu$. Volume was normalized by the volume obtained with the lowest $\mu$ (4.3e-13, 2.4e-12, and 2.1e-38 in particular).}
    \label{fig:pareto}
\end{figure*}

\subsection{GoTube can trade runtime for reachtube tightness}
In our last experiment, we introduced a new set of benchmark models entirely based on continuous-time recurrent neural networks. 
The first model is an unstable linear dynamical system of the form $\dot{x} = Ax + Bu$ that is stabilized by a CT-RNN policy via actions $u$. The second model corresponds to the inverted pendulum environment, which is similar to the CartPole environment but differs in that the control actions are applied via a torque vector on the pendulum directly instead of moving a cart. The CT-RNN policies for these two environments were trained using deep RL.
Our third new benchmark model concerns the analysis of the learned dynamics of a CT-RNN trained on supervised data. In particular, by using the reachability frameworks, we aim to assess if the learned network expressed oscillatory behavior. The CT-RNN state vector consists of 16 dimensions, which is twice as much as existing CT-RNN reachability benchmarks \cite{gruenbacher2020lagrangian}. 

Here, we study how GoTube can trade runtime for the volume of the constructed reachtube through its tightness factor $\mu$.
In particular, we run GoTube on our newly proposed benchmark with various values of $\mu$. We then plot GoTube's runtime (x-axis) and volume size (y-axis) as a function of $\mu$. The resulting curves show the Pareto-front of runtime-volume trade-off achievable with GoTube.

Figure \ref{fig:pareto} shows the results for a time horizon of 10s in the first two examples, and of 2s in the last example.
Our results demonstrate that GoTube can adapt to different runtime and tightness constraints and set a new benchmark for future methods to compare with.

\section{Discussions, Scope and Conclusions}

We proposed GoTube, a new stochastic verification algorithm that provides robustness guarantees (also safety guarantees if a set of states to be avoided is given)  for high-dimensional, time-continuous systems. GoTube is stable and sets the state-of-the-art in terms of its ability to scale to time horizons well beyond what has been previously possible. It also allows a larger perturbation radius for the initial ball, for which other verification methods fail. Lastly, GoTube's scalability enables it to readily handle the verification of advanced continuous-depth neural models, a setting where state-of-the-art deterministic approaches fail. 

\noindent\textbf{SLR versus GoTube?} SLR combines symbolic with statistical reachability techniques. However, no implementation is available to date. For comparison purposes, we implemented SLR on our own and observed that while it does not blow up in space, it blows up in time. As a consequence, we were not able to use SLR to construct reachtubes for our high-dimensional benchmarks.

\noindent\textbf{Sample blow up in GoTube?}
As a pure Monte-Carlo technique, the number of samples $N$ to be taken depends on both the confidence coefficient $\lambda$ and the tightness coefficient $\mu$ as well as on the system's dimensionality. As a consequence, for very small values of these coefficients, the number of samples tends to blow up. The goal of symbolic techniques is exactly the one to avoid such a blowup. However, in our experiments, we observed that GoTube outperformed in all cases the symbolic techniques. This implies that the overapproximation error of symbolic techniques is more problematic than the blowup in the number of samples for a large number of dimensions.

\noindent\textbf{What about Gaussian Processes?} Gaussian Processes (GPs) are powerful stochastic models which can be used for stochastic reachability analysis~\cite{gp} and uncertainty estimation for stochastic dynamical systems~\cite{gal2016uncertainty}. The major shortcoming of GPs is that they simply cannot scale to the complex continuous-time systems that we tested here. Moreover, Gaussian Processes have a large number of hyperparameters, which can be challenging to tune across different benchmarks. 

\noindent\textbf{Limitations of GoTube.} GoTube does not necessarily perform better in terms of average volume of the bounding balls for smaller tasks and shorter time horizons if not choosing a very small $\mu$, as shown in Table~\ref{tab:volume}. GoTube is not yet suitable for the verification of stochastic dynamical systems, for instance, Neural Stochastic Differential Equations (Neural SDEs) \cite{li2020scalable,xu2021infinitely}. Although GoTube is considerably more computationally efficient than existing methods, the dimensionality of the system, as well as the type of numerical ODE solver exponentially, affect their performance. We can improve on this limitation by using Hypersolvers \cite{poli2020hypersolvers}, closed-form continuous depth models, and compressed representations of neural ODEs.

\noindent\textbf{Future of GoTube.} GoTube opens many avenues for future research. The most straightforward next step is to search for better intermediate steps in Algorithm \ref{algorithm:GoTube}. For instance, better ways to compute the Lipschitz constant and to improve the sampling process. GoTube is now applicable for complex deterministic ODE systems; it would be an important line of work to find ways to marry reachability analysis with machine learning approaches to verify neural SDEs as well. 
Last but not least, we believe that there is a close relationship between stochastic reachability analysis and uncertainty estimation techniques used for deep learning models \cite{abdar2021review}. Uncertainty-aware verification could be worth exploring based on what we learned with GoTube. 


\bibliography{aaai22}

\clearpage
\setcounter{theorem}{0}

\input{supplements}


\end{document}

%% file: supplements.tex
\newcommand{\beginsupplement}{%
        \setcounter{table}{0}
        \renewcommand{\thetable}{S\arabic{table}}%
        \setcounter{algorithm}{0}
        \renewcommand{\thealgorithm}{S\arabic{algorithm}}%
        \setcounter{equation}{0}
        \renewcommand{\theequation}{S\arabic{equation}}%
        \setcounter{figure}{0}
        \renewcommand{\thefigure}{S\arabic{figure}}%
        \setcounter{section}{0}
        \renewcommand{\thesection}{S\arabic{section}}%
     }

\beginsupplement

\section{Appendix\\ {\large Proofs of the Theorems}}
\begin{lemma}[Stochastic lower bound $F_{L,\gamma}$]\label{app_lemma:F_L}
    Consider the experiment of randomly sampling two times $m$ points of the initial ball $\calB_0$: $(a_1,\dots,a_m)$ and $(b_1,\dots,b_m)$.
    Let $g\,:\,\R^n\,\rightarrow\,\R$ be a real-valued function and $X\,{=}\,\max_{i=1}^m|g(a_i)\,{-}\,g(b_i)|/\|a_i\,{-}\,b_i\|$ be a random variable with the unknown cumulative distribution function $F$. Let $(x_1, \dots, x_n)\sim X$ be independent, identically distributed samples with the empirical distribution function $\hat{F}_n(x)\,=\,\sum_{i=1}^n\mathbbm{1}_{x_i\le x}$. Let $G_n$ be a generalized extreme value distribution fitted to the empirical distribution function $\hat{F}_n$ and let $D_n^-$ describe the goodness of fit, being the one-sided Kolmogorov–Smirnov statistic:
    \begin{align}\label{app_eq:ks-statistic}
        D_n^- \,=\,\sup_x (G_n(x) - \hat{F}_n(x))
    \end{align}
    Given the confidence level $\gamma$ and $\alpha = \min(\gamma, 0.5)$, then let us define $\epsilon_{n,\gamma}$ and $F_{L,\gamma}$ as follows:
    \begin{align}
        \epsilon_{n,\gamma} &= \sqrt{\frac{\ln{\frac1{\alpha}}}{2n}}\label{app_eq:dkw_epsilon}\\
        F_{L,\gamma}(x) &= G_n(x) - \epsilon_{n,\gamma} - D_n^-\label{app_eq:F_L bound}
    \end{align}
    Then it holds that:
    \begin{align}\label{app_eq:prob F_L bound}
        Pr(\sup_x(F_{L,\gamma}(x)-F(x))\le 0)\ge 1-\gamma,
    \end{align}
    which intuitively means that $F_{L,\gamma}$ is a lower bound of $F$ with confidence $\gamma$.
\end{lemma}
\emph{Proof.} The Fisher-Tippett-Gnedenko theorem states that the distribution of a normalized maximum converges to the generalized extreme value distribution, if the distribution of the normalized maximum does converge. So intuitively that theorem is similar to the central limit theorem for the averages, but for the normalized maxima. Consequently, we start by fitting the empirical distribution function $\hat{F}_n$ by a generalized extreme value distribution $G_n$ and compute Eq.~\eqref{app_eq:ks-statistic}.

The Dvoretzky-Kiefer-Wolfowitz inequality~\cite{dkw1956} with a tight constant determined by~\cite{massart1990}, states that for all $\epsilon\,\ge\,\sqrt{\frac1{2n}\ln2}$, it holds that:
\begin{align}
    Pr(\sup_x(\hat{F}_n(x)-F(x))>\epsilon)\le e^{-2n\epsilon^2}
\end{align}
Solving $\gamma = e^{-2n\epsilon^2}$ for $\epsilon$ and considering Massarts lower bound for $\epsilon$, yields:
\begin{align}\label{app_eq:DKW inequality}
    \begin{split}
        &Pr(\sup_x(\hat{F}_n(x)-F(x))>\epsilon_{n,\gamma}) \le \gamma,
    \end{split}
\end{align}
with $\epsilon_{n,\gamma}$ as defined in Eq.~\eqref{app_eq:dkw_epsilon}. We use the triangular inequality for supremum and the monotony of the probability measure as follows:
\begin{align}
    &Pr\big(\sup_x\big(G_n(x)-F(x)\big)>\epsilon_{n,\gamma} + D_n^-\big) =\\
    &= Pr\Big(\sup_x\big(G_n(x)-\hat{F}_n(x)+\\
    &\qquad + \hat{F}_n(x)-F(x)\big)>\epsilon_{n,\gamma} + D_n^-\Big)\\
    &\le Pr\Big(\sup_x\big(G_n(x)-\hat{F}_n(x)\big) + \\
    &\qquad +\sup_x\big(\hat{F}_n(x)-F(x)\big)>\epsilon_{n,\gamma} + D_n^-\Big)\\
    &\overset{\eqref{app_eq:ks-statistic}}{=} Pr\big(\sup_x\big(\hat{F}_n(x)-F(x)\big)>\epsilon_{n,\gamma}\big)\\
    & \overset{\eqref{app_eq:DKW inequality}}{\le} \gamma,
\end{align}
from which it follows directly, that Eq.~\eqref{app_eq:prob F_L bound} hold.
\begin{figure}[h]
    \centering
    \includegraphics[width=0.48\textwidth]{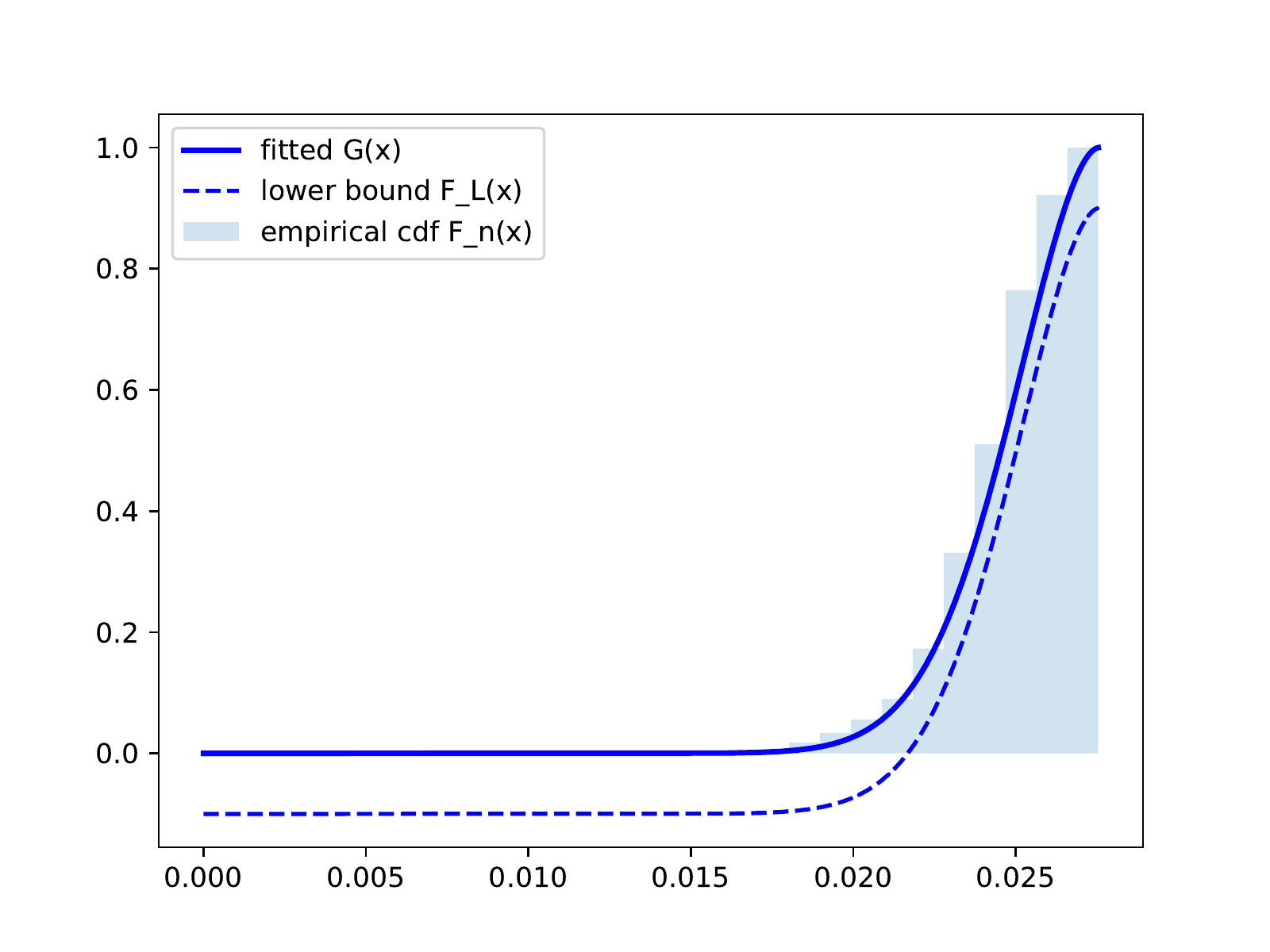}
    \caption{Visualisation of the stochastic lower bound $F_{L,\gamma}$ of Lemma~\ref{app_lemma:F_L}.}
    \label{app_fig:lemma1}
\end{figure}
\begin{theorem}[Radius of Stochastic Lipschitz Caps]\label{app_thm:lipschitz cap}
    Given a continuous-depth model $f$ from Eq.~(1) in the main paper ($\partial_t x\,{=}\,f(x)$ with $x(t_0)\,{\in}\,\,B(x_0, \rd_0)$), $\gamma\,{\in}\,(0,1)$, $\mu\,{>}\,1$, target time $t_j$,
    the set of all sampled points $\calV$, the number of sampled points $N\,{=}\,|\calV|$, the sample maximum $\bar{m}_{j,\calV}\,{=}\,\max_{x\in\calV} d_j(x)$, the IVP solutions $\chi(t_j,x)$, and the corresponding stretching factors $\lambda_x\,{=}\,\|\partial_x\chi(t_j,x)\|$ for all $x\,{\in}\,\calV$. Let us define $\hat{\gamma} = 1-\sqrt{1-\gamma}$.
    Let $\Delta\lambda_{\calV}$ be the $\sqrt{1-\gamma}$-quantile of a stochastic lower bound $F_{L,\hat{\gamma}}$ as defined in Eq.~\eqref{app_eq:F_L bound} of Lemma~\ref{app_lemma:F_L}:
    \begin{align}\label{app_eq:expected difference quotient}
        \Delta\lambda_{\calV}(\gamma) = F_{L,\hat{\gamma}}^{-1}(\sqrt{1-\gamma}),
    \end{align}
    Let $r_x$ be defined as:
    \begin{align}\label{app_eq:cap radius}
        r_{x} = \frac{\left(-\lambda_x + \sqrt{\lambda_x^2 + 4\cdot\Delta\lambda_{x,\calV}\cdot(\mu\cdot\bar{m}_{j,\calV}-d_j(x))}\right)}{2\cdot\Delta\lambda_{x,\calV}},
    \end{align}
    then it holds that:
    \begin{align}\label{app_eq:cap probability}
        \Pr\left(d_j(y) \le \mu\cdot \bar{m}_{j,\calV}\right)\ge 1-\gamma\quad \forall y\in B(x,r_x)^S,
    \end{align}
    and thus that $B(x, r_x)^S$ is a $\gamma, t_j$-Lipschitz cap.
\end{theorem}
\emph{Proof.} Let $\{x_1,\dots,x_n\}$ be $n$ independent experiments by sampling from $X\,{=}\,\max_{i=1}^m|\lambda_{a_i}\,{-}\,\lambda_{b_i}|/\|a_i\,{-}\,b_i\|$ as defined in Lemma~\ref{app_lemma:F_L}, where each variable is the maximum of $m$ executions. 
From Eq.~\eqref{app_eq:prob F_L bound} it follows that:
\begin{align}\label{app_eq:cdf bound}
    Pr(F_{L,\hat{\gamma}(x)} \le F(x))\ge 1-\hat{\gamma} = \sqrt{1-\gamma}
\end{align}
Let us now derive the probability of $X$ being less or equal to $\Delta\lambda_{\calV}$ defined by Eq.~\eqref{app_eq:expected difference quotient}. 
For any sets $A, B$ it holds that $\Pr(A)\ge \Pr(A\cap B) = \Pr(A | B) \cdot \Pr(B)$, thus:
\begin{align}
    &Pr(X \le \Delta\lambda_{\calV}) \ge\\
    &\ge Pr\Big(X\le\Delta\lambda_{\calV} | F_{L,\hat{\gamma}(x)} \le F(x)\Big)\cdot\label{app_eq:quantile1}\\
    &\qquad\cdot Pr\Big(F_{L,\hat{\gamma}(x)} \le F(x)\Big)
\end{align}
Let us have a look on Eq.~\eqref{app_eq:quantile1}: As $Pr(X\le \Delta\lambda)=F(\Delta\lambda)$ and we are looking for the conditional probability depending on $F_{L,\hat{\gamma}(x)} \le F(x)$, we can use $F_{L,\hat{\gamma}}(\Delta\lambda)$ as a lower bound of Eq.~\eqref{app_eq:quantile1} and thus, using Eq.~\eqref{app_eq:cdf bound}:
\begin{align}
    Pr(X\le \Delta\lambda_{\calV}) \ge F_{L,\hat{\gamma}}(\Delta\lambda_{\calV})\cdot \sqrt{1-\gamma}
\end{align}
As Eq.~\eqref{app_eq:expected difference quotient} defines $\Delta\lambda_{\calV}$ as the $\sqrt{1-\gamma}$-quantile of $F_{L,\gamma}$, we can further state that
\begin{align}\label{app_eq:confidence lipschitz max}
    \begin{split}
    \Pr(X \le \Delta\lambda_{\calV})\ge 1-\gamma, \\\quad \textrm{with}\quad X=\max_{i=1}^m\left[\frac{|\lambda_{a_i} -  \lambda_{b_i}|}{\|a_i-b_i\|}\right]
    \end{split}
\end{align}
Let $Y\,=\, |\lambda_a-\lambda_b|/\|a-b\|$ be another random variable with $a, b$ being to sample points of the initial ball $\calB_0$. From Eq.~\eqref{app_eq:confidence lipschitz max} it holds that
\begin{align}\label{app_eq:confidence lipschitz}
    Pr(Y\le \Delta\lambda_\calV)\ge Pr(X\le \Delta\lambda_\calV)\ge 1-\gamma
\end{align}
It trivially holds for $x, y \in \calV$ that:
\begin{align}
    \lambda_y &= \lambda_x + \frac{\lambda_y - \lambda_x}{\|x-y\|}\cdot \|x-y\|\nonumber\\
    &\le \lambda_x + \frac{|\lambda_x - \lambda_y|}{\|x-y\|}\cdot \|x-y\|
\end{align}
From Eq.~\eqref{app_eq:confidence lipschitz} it follows that:
\begin{align}
    &Pr\Bigg(\lambda_x + \frac{|\lambda_x - \lambda_y|}{\|x-y\|}\cdot \|x-y\| \le \\
    &\qquad\le \lambda_x + \Delta\lambda_{x,\calV}\cdot \|x-y\|\Bigg)\ge 1-\gamma
\end{align}
and using the monotony of the probability measure:
\begin{align}
    \Pr\big(\lambda_y &\le \lambda_x + \Delta\lambda_{x,\calV}\cdot \|x-y\|\big) \ge 1-\gamma\label{app_eq:confidence bound}
\end{align}
Using the mean value inequality for vector-valued functions it holds that:
\begin{align*}
    & |d_j(x) - d_j(y)|= | \left\lVert \chi(t_j,x) - \chi(t_j,x_0)\right\rVert -\\ &\quad- \left\lVert \chi(t_j,y) - \chi(t_j,x_0)\right\rVert|\quad \textrm{\{triangle inequality\}}\\
    &\quad \le \|\chi(t_j,x) - \chi(t_j,y)\|\quad \textrm{\{mean value theorem\}}\\
    &\Rightarrow\exists z\in [x,y] \colon  |d_j(x) - d_j(y)|\\ &\quad\le \|\partial_x \chi(t_j,z) \| \|x-y\| = \lambda_z \cdot \|x-y\|
\end{align*}
Combining this with Eq.~\eqref{app_eq:confidence bound} and thus using $\lambda_x + \Delta\lambda_{x,\calV}\cdot \|x-y\|$ as a probabilistic upper bound for $\lambda_z$, we obtain the following results for all $y$ with $\|x-y\| \le r_x$:
\begin{align}
    &\Pr\big(|d_j(x)-d_j(y)|\le\nonumber\\
    &\quad\le (\lambda_x + \Delta\lambda_{x,\calV}\cdot \|x-y\|) \cdot \|x-y\|\big) \ge 1-\gamma \nonumber\\
    \begin{split}
    &\Pr\big(|d_j(x)-d_j(y)|\le\\
    &\quad\le (\lambda_x + \Delta\lambda_{x,\calV}\cdot r_x) \cdot r_x\big) \ge 1-\gamma
    \end{split}
\end{align}
As $r_x$ defined like in Eq.~\eqref{app_eq:cap radius} is the solution of the quadratic equation $\mu\cdot\bar{m}_{j,\calV}-d_j(x)=\lambda_x r_x + \Delta\lambda_{x,\calV} r_x^2$, it holds that:
\begin{align}
\begin{split}
    &\Pr\big(|d_j(x) - d_j(y)| \le \mu\cdot\bar{m}_{j,\calV}-d_j(x))\big) \ge\\
    &\quad \ge 1-\gamma \quad\forall y\in B(x, r_x)^S\label{app_eq:probabilistic lipschitz}
\end{split}
\end{align}
We now distinguish between two cases for $y$: (a)~$d_j(y)\le d_j(x)$ and (b)~$d_j(y) \ge d_j(x)$. In case (a) it is trivial: $d_j(y) \le d_j(x) \le \mu \cdot \bar{m}_{j,\calV}$. Having case (b), Eq.~\eqref{app_eq:probabilistic lipschitz} is equivalent to
\begin{align}
    \Pr\big(d_j(y) - d_j(x) &\le \mu\cdot\bar{m}_{j,\calV}-d_j(x))\big) \ge 1-\gamma\nonumber\\ &\Longleftrightarrow\nonumber\\
    \Pr\big(d_j(y) &\le \mu\cdot\bar{m}_{j,\calV})\big) \ge 1-\gamma,
\end{align}
thus Eq.~\eqref{app_eq:cap probability} holds and $B(x,r_x)^S$ is a Lipschitz cap.

\begin{theorem}[Convergence via Lipschitz Caps]\label{app_thm:stochastic guarantee} Given the tightness factor $\mu > 1$, the set of all sampled points $\calV$ and the sample maximum $\bar{m}_{j,\calV} = \max_{x\in\calV} d_j(x)$. Let the initial ball maximum be defined by $m^\star_j=\max_{x\in\calB_0} d_j(x)$. Then:
    \begin{align}\label{app_eq:stochastic guarantee}
        \hspace*{-2ex}\forall\gamma\in(0,1),\exists N\in\mathbb{N}\textrm{ s.t. }
        \Pr(\mu\cdot\bar{m}_{j,\calV}\ge m^\star_j) \ge 1-\gamma
    \end{align}
    where $N=|\calV|$ is the number of sampled points.
\end{theorem}
\emph{Proof.} Let $x^\star_j$ be a point such that $d_j(x^\star_j) = m^\star_j$. Given $\gamma\in(0,1)$ and cap radii $r_x$ as defined in Eq.~\eqref{app_eq:cap radius}, we know from the definition of a spherical cap that
\begin{align}
    &p_{r_x} = \Pr(B(x, r_x)^S\owns x_j^\star) = \frac{\area(B(x, r_x)^S)}{\area{(\calB_0})}\\
    &\textrm{and thus it holds that:}\nonumber\\
    &\Pr(\exists y\in\calV\colon B(y, r_y)^S\owns x_j^\star) = 1 - \prod_{x\in\calV} \left(
        1 - p_{r_x}
        \right)
\end{align}
We derive a lower bound of $r_x$ by using the first sample $x_{j,1}$ and replacing the values in Eq.~\eqref{app_eq:cap radius} as follows:
\begin{align}
    &\mu\cdot \bar{m}_{j,\calV} - d_j(x) \\
    &\quad\ge \mu\cdot \bar{m}_{j,\calV} - \bar{m}_{j,\calV} = (\mu - 1)\cdot \bar{m}_{j,\calV} \\
    &\quad\ge (\mu - 1) \cdot d_j(x_{j,1}),
\end{align}
thus a lower bound of all Lipschitz cap radii is given by
\begin{align}
    & r_{bound} = \nonumber\\
    &\quad=\frac{-\lambda_x + \sqrt{\lambda_x^2 + 4\cdot\Delta\lambda_{x,\calV}\cdot(\mu - 1) \cdot d_j(x_{j,1})}}{2\cdot\Delta\lambda_{x,\calV}} \le \nonumber\\
    &\quad\le r_x \quad\forall x\in \calV\nonumber\\
    \begin{split}
    &\Rightarrow \Pr(\exists y\in\calV\colon B(y, r_y)^S\owns x_j^\star) \ge \\
    &\quad\ge 1 - \left(
        1 - p_{r_{bound}}
        \right)^N\label{app_eq:lower bound probability}
    \end{split}
\end{align}
As in the limit of $N\rightarrow \infty$ the probability of Eq.~\eqref{app_eq:lower bound probability} is 1, it follows that $\forall \gamma\in (0,1)~\exists N \in\mathbb{N}\colon\Pr(\exists x\in\calV\colon\,B(x,r_x)^S\owns x^\star_j)\ge\sqrt{1-\gamma}$.

Using a set of sampled points $\calV$ with cardinality $N$ and using $\hat{\gamma} = 1-\sqrt{1-\gamma}$ as the error rate for the upper bound $\Delta\lambda_x$ of the confidence interval in Eq.~\eqref{app_eq:expected difference quotient}. Using the result of Theorem~\ref{app_thm:lipschitz cap}, the resulting probability $\forall y\in B(x,r_x)^S$ is:
\begin{align}\label{app_eq:proof lipschitz cap}
    \Pr\left(d_j(y) \le \mu\cdot \bar{m}_{j,\calV}\right)\ge 1-\hat{\gamma} = \sqrt{1-\gamma}
\end{align}
If there is an $x\in\calV$ such that $B(x,r_x)^S\owns x_j^\star$, then Eq.~\eqref{app_eq:proof lipschitz cap} obviously holds also for $x_j^\star$, thus:
\begin{align*}
    \Pr( d_j(x^\star) \le \mu\cdot\bar{m}_{j,\calV} |\exists x\in\calV\colon B(x,r_x)^S\owns x^\star)\ge \sqrt{1-\gamma}
\end{align*}
For any sets $A, B$ it holds that $\Pr(A)\ge \Pr(A\cap B) = \Pr(A | B) \cdot \Pr(B)$, and using:
\begin{align}
    A &= (\mu\cdot\bar{m}_{j,\calV}\ge m^\star_j)\\
    B &= (\exists x\in\calV\colon B(x,r_x)^S\owns x_j^\star)
\end{align}
it follows that $\Pr(\mu\cdot\bar{m}_{j,\calV}\ge m^\star_j)\ge \Pr(A|B) \cdot \Pr(B) = 1-\gamma$ and therefore Eq.~\eqref{app_eq:stochastic guarantee} holds.
